\documentclass[a4paper,fleqn]{cas-dc}

\usepackage[numbers]{natbib}

\def\tsc#1{\csdef{#1}{\textsc{\lowercase{#1}}\xspace}}
\tsc{WGM}
\tsc{QE}
\tsc{EP}
\tsc{PMS}
\tsc{BEC}
\tsc{DE}

\begin{document}
\let\WriteBookmarks\relax
\def\floatpagepagefraction{1}
\def\textpagefraction{.001}
\shorttitle{Multi-Attention Fusion Drowsy Driving Detection Model}
\shortauthors{Shulei QU et~al.}

\title [mode = title]{Multi-Attention Fusion Drowsy Driving Detection Model}

\author[1]{Shulei QU}
\author[1]{Zhenguo Gao}
\cormark[1]
\credit{supervision, writing and editing}
\author[1]{Xiaoxiao Wu}
\credit{writing and reviewing}
\author[1]{Yuanyuan Qiu}
\credit{writing and reviewing}
\address[1]{School of Computer Science and Technology, Huaqiao University, China}
\cortext[cor1]{Corresponding author}

\begin{abstract}
Drowsy driving represents a major contributor to traffic accidents, and the implementation of driver drowsy driving detection systems has been proven to significantly reduce the occurrence of such accidents. Despite the development of numerous drowsy driving detection algorithms, many of them impose specific prerequisites such as the availability of complete facial images, optimal lighting conditions, and the use of RGB images. In our study, we introduce a novel approach called the Multi-Attention Fusion Drowsy Driving Detection Model (MAF). MAF is aimed at significantly enhancing classification performance, especially in scenarios involving partial facial occlusion and low lighting conditions. It accomplishes this by capitalizing on the local feature extraction capabilities provided by multi-attention fusion, thereby enhancing the algorithm's overall robustness. To enhance our dataset, we collected real-world data that includes both occluded and unoccluded faces captured under nighttime and daytime lighting conditions. We conducted a comprehensive series of experiments using both publicly available datasets and our self-built data. The results of these experiments demonstrate that our proposed model achieves an impressive driver drowsiness detection accuracy of 96.8\%.

\end{abstract}


\maketitle

\section{Introduction}

Drowsy driving detection is a critical function of Advanced Driver Assistance Systems (ADAS) aimed at preventing accidents and injuries caused by drowsy driving, which can affect both drivers and pedestrians. According to a survey, as of the end of September 2022, the number of motor vehicles in the country reached 412 million, with 76\% (315 million) being automobiles. The total number of motor vehicle drivers reached 499 million, with 461 million being automobile drivers.

Data from the World Health Organization reveals that approximately 1.2 million people worldwide lose their lives in car accidents every year, with one person killed in a traffic accident every 25 seconds. In today's rapidly evolving socio-economic landscape, people's lives are becoming increasingly fast-paced, making drowsy driving one of the most significant factors affecting road safety. Research conducted by the Canadian TIRF organization found that over half (57\%) of surveyed drivers consider driving while fatigued to be a highly dangerous behavior, yet one in five (20\%) admitted to having done so at least once\cite{1beirness2005road}. The German Transportation Safety Commission discovered through a survey that approximately 20\% of car accidents on highways are attributed to driver drowsiness.

According to statistics from China's Ministry of Transportation and Communications (MOTC) in 2010, a total of 1,890 road traffic accidents were caused by driver drowsiness. Unlike other dangerous driving behaviors such as driving under the influence of alcohol and speeding, drowsy driving is often inconspicuous and more challenging to detect and manage. As a result, it has emerged as a key factor in traffic accidents. To mitigate the associated loss of lives and property, the development of high-performance drowsy driving detection  methods is imperative. This is particularly crucial for drivers engaged in passenger transportation, freight transportation, and other operations, where extended periods of continuous driving are common due to occupational requirements.

The most commonly used and effective methods are those based on the percentage of eye closure (PERCLOS) within a predefined time window. Some existing PERCLOS-based methods typically employ advanced facial key point detection models to extract eye key points and calculate the degree of eye closure based on these key points. However, such methods often require manual selection of a suitable PERCLOS threshold, and even a slight shift in the key point coordinates can lead to erroneous eye closure calculations. Additionally, the presence of obscured faces can render key points extraction entirely ineffective.

Recently, Sanghyuk Park et al.\cite{31park2016driver} designed an end-to-end convolutional neural network that combines multiple classification networks to detect fatigue. While this approach alleviates some of the limitations of PERCLOS-based methods, the network solely relies on general classification networks and does not incorporate specific designs for fatigue detection.

In this paper, we introduce the Multi-Attention Fusion Drowsy Driving Detection Model(MAF). While Transformers\cite{4vaswani2017attention} have demonstrated remarkable performance in various visual tasks, including image classification\cite{5dosovitskiy2020image}\cite{6heo2021rethinking}\cite{7meng2022adavit}, object detection\cite{8carion2020end}\cite{9zhu2020deformable}, video classification\cite{be}\cite{11liu2022video}\cite{12wang2022efficient}\cite{13wang2022bevt}, their computational demands can be substantial. To address this concern, MAF is constructed by combining Convolutional Neural Networks and Transformers.

The process begins with the extraction of a feature map from the input image through a backbone network's basic feature extraction. This feature map undergoes convolutional attention processing via multiple parallel LANets, enhancing regions of interest for drowsiness detection. We employ a random attention map dropout method to compel the model to explore additional feature regions while avoiding overreliance on any specific region that might lead to model failure. Additionally, we introduce a set of learnable patches to explicitly represent local features, complementing the extracted feature patches. These learnable patches interact with the feature patches through cross-attention to further enhance effective regions.

Through iterative processing, a global feature representation is derived for predicting whether a driver is drowsy or alert. Our model is trainable in an end-to-end fashion. Extensive experiments were conducted on both self-collected data and the RLDD dataset\cite{14ghoddoosian2019realistic}. The results demonstrate the superior performance of our approach. In summary, our work contributes in the following ways:
\begin{itemize}
    \item We introduce MAF, an end-to-end driver drowsiness detection model, that combines multiple attention mechanisms.
    \item We employ random dropout and incorporate learnable patches to enhance the model's capacity for modeling local features.
    \item We conducted extensive experiments to validate the effectiveness of our methodology.
\end{itemize}

\begin{figure*}[]
    \centering
    \includegraphics[width=1\linewidth]{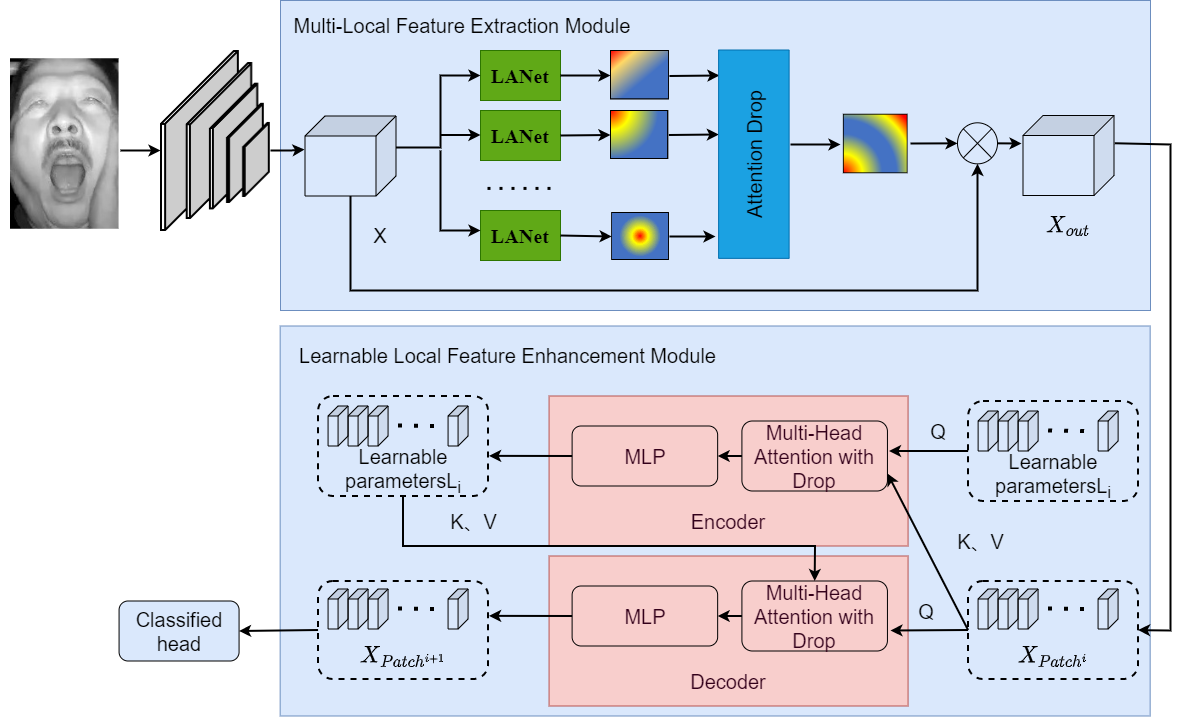}
    \caption{Network framework of Multi-Attention Fusion Drowsy Driving Detection Model (MAF)}
    \label{fig:1}
\end{figure*}

\section{Related Work}

In this section, we briefly introduce driver drowsiness detection methods in different directions based on the analyzed features. According to the characteristics of the detection object and the methods used, driver drowsiness detection is subdivided into detection methods based on driver physiological signals, detection methods based on driver behavioral characteristics, detection methods based on vehicle characteristics, and detection methods based on information fusion.

\textbf{The detection methods based on the driver's physiology.} Drowsiness detection based on physiological features primarily involves the extraction of physiological signals from the driver's body, with subsequent assessments of the driver's drowsiness based on changes in these physiological signals. Commonly employed physiological signals in this method encompass electroencephalogram (EEG), electrocardiogram (ECG), electromyogram (EMG), and others. EEG signals, for example, can be decomposed into four rhythms ($\alpha$-wave, $\beta$-wave, $\theta$-wave, $\delta$-wave). In cases of driver fatigue, the energy of $\delta$ and $\theta$ waves increases while the energy of $\alpha$ and $\beta$ waves decreases. This shift can be utilized to determine the level of fatigue\cite{15fu2016dynamic}. Ibtissem Belakhdar et al.\cite{16belakhdar2016detecting} introduced a single-channel automated drowsy driving detection method based on Artificial Neural Networks (ANNs). They utilized the Fast Fourier Transform to compute nine features from the EEG channels. After integrating these features into the ANN classifier, they achieved classification accuracies of 86.1\% for detecting drowsiness and 84.3\% for identifying sobriety. S.S. Poorna et al.\cite{17poorna2018drowsiness} proposed a technique involving principal component analysis on EEG data to extract dominant eye impulses. This method generated two sets of data: one set of feature vectors that represented blinks and another that excluded blinks. Both KNN and ANN classifiers were employed to classify these features and recognize whether the driver was awake or drowsy, achieving accuracies of 80\% and 85\%, respectively.

\textbf{The detection methods based on vehicle characteristics.} The detection method based on vehicle characteristics involves monitoring various data during vehicle operation, including parameters such as driving speed, steering angle, lane offset, and more. This method focuses on analyzing the multivariate correlations between these data and their changes during driving, comparing them with data from typical, normal driving conditions. By assessing the discrepancies between the two sets of data, this approach can judge the appropriateness of driving behavior and reveal abnormal driving patterns as well as the driver's level of fatigue. For instance, Li et al.\cite{18li2017online} collected steering wheel angle data (SWA) using a sensor mounted on the steering column and extracted valuable insights from the SWA data, enabling online driver drowsiness detection. Wang et al.\cite{19wang2016drowsy} combined steering wheel angle information with data on vehicle longitudinal and lateral acceleration, utilizing various time window sizes for training. They employed the Random Forest algorithm to analyze driver vehicle parameter data in both fatigue and normal states, ultimately determining the driver's level of drowsiness.

\textbf{The detection methods based on driver behavioral characteristics.} When a driver experiences fatigue, several noticeable features emerge, including left and right eye drifting, a gradual reduction in the degree of eye opening, increased blink frequency, yawning, and frequent head nodding\cite{20ngxande2017driver}. The method based on driver behavioral features aims to assess a driver's fatigue by capturing image information of the driver, extracting relevant features, and analyzing them. Presently, available features for this approach mainly encompass eye features, mouth features, head posture features, facial expression features, and more. Bhargava Reddy et al.\cite{34reddy2017real} proposed a 2-stream network, with each stream extracting image features from different parts of the left eye and mouth. They fused the features extracted from these four streams to produce detection results and achieved an accuracy of 91.3\% on a dataset collected by the authors themselves. Mohit Dua et al.\cite{32dua2021deep} proposed a drowsiness detection system. The system employs four deep learning models (AlexNet, VGG-FaceNet, FlowImageNet, and ResNet) to analyze various driver features, including hand gestures, facial expressions, behavioral features, and head movements. These models classify drivers into four categories: non-drowsiness, drowsiness with eye blinking, yawning, and nodding. The system achieves an 85\% accuracy rate.

\textbf{The detection methods based on information fusion.} The detection method based on information fusion involves the integration of multiple features derived from driver physiological parameters, behavioral characteristics, and vehicle driving information. Information fusion-related theories are employed to detect driver fatigue effectively. For instance, in a study by Jasper Gielen and Jean-Marie Aerts\cite{23gielen2019feature}, nose and wrist temperatures, as well as heart rates of 19 participants in a driving simulation, were monitored. The integration of these feature variables yielded a classification accuracy of 89.5\%. Similarly, Amirudin et al.\cite{24amirudin2018detection} combined physiological parameters and behavioral features for drowsiness detection. They extracted $\delta$, $\theta$, and $\alpha$ wave features from EEG signals, as well as eye features from video sequences. These EEG features were classified using KNN, ANN, and SVM, and the Viola-Jones algorithm was employed to detect the driver's eye state. Ultimately, the $\theta$ wave physiological features were used in combination with the eye state to detect the driver's overall state.

\section{Method}

We propose the Multi-Attention Fusion Drowsy Driving Detection Model(MAF). Fig. \ref{fig:1} shows the overall architecture of MAF, which contains a backbone network, a multi-local feature extraction module, and a learnable local feature enhancement module. Each module is described separately in the following.

\subsection{Multi-Local Feature Extraction Module}

To encourage the neural network to extract local features that are effective for drowsiness detection, we introduce the Multiple Local Feature Extraction Module. This module not only prioritizes the most prominent feature but also captures various other local features essential for classification. It comprises several convolutional attention layers and an attention map random dropout layer.

\begin{figure}[htbp]
    \centering
    \includegraphics[width=0.8\linewidth]{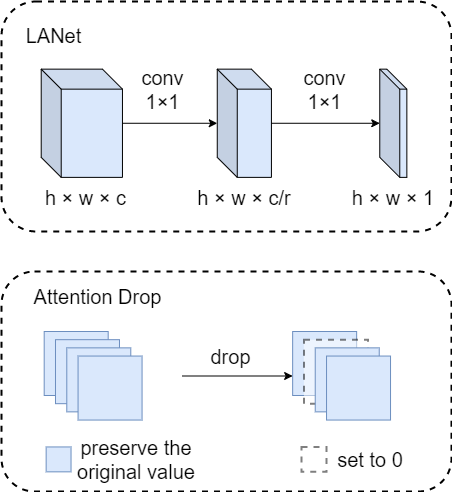}
    \caption{The network architecture of LANet and the working mechanism of Attention Drop}
    \label{fig:lanet_drop}
\end{figure}

We incorporate LANet\cite{25wang2019ls}, a spatial attention network designed to automatically localize crucial facial regions, in parallel to extract multiple local features. As depicted in Fig. \ref{fig:lanet_drop}, LANet consists of two 1 × 1 convolutional layers. Given the feature map from the backbone network as $X\in\mathbb{R}^{\mathcal{C}\times H\times W}$, where c represents the number of feature channels. The first 1 × 1 convolutional layer reduces the feature channels to ${c/r}$ after which a ReLU layer follows. Here, r is a hyperparameter defining the compression rate of the feature channels. Subsequently, a second 1 × 1 convolutional layer further reduces the feature channels to 1, followed by a sigmoid activation function to produce the attention map. Multiple attention maps are generated by multiple parallel LANets, and they are concatenated along the channel dimension.

The subsequent attention graph random dropout layer randomly discards one of the attention maps with a probability of $p$, setting the value of the discarded attention map to 0. This process, as illustrated in Fig. \ref{fig:lanet_drop}, ensures that the model focuses on a broader range of local feature regions likely to be effective, rather than solely concentrating on the most dominant features. The remaining multiple attention maps undergo maximum pooling in the channel dimension to produce the final attention map. Element-wise multiplication of the attention maps with the feature maps from the backbone network yields the final feature maps.

\subsection{Learnable Local Feature Enhancement Module}
The Learnable Local Feature Enhancement Module has the capability to explicitly learn multiple local features within the feature maps. To achieve this, we introduce a set of learnable patches $L\in\mathbb{R}^{N\times C}$, where N is a hyperparameter that governs the maximum number of local features to be learned. In MAF, this parameter is set to be equal to the number of LANets in parallel with those in the multi-local feature extraction module.

\begin{figure}[h]
    \centering
    \includegraphics[width=0.6\linewidth]{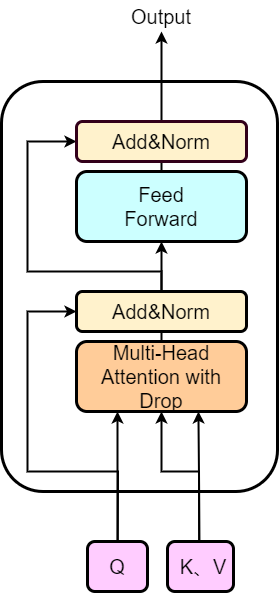}
    \caption{The diagram represents a single attention unit. The Encoder and Decoder share the same structure, with multiple of these units stacked together}
    \label{fig:coder}
\end{figure}

Each unit within the learnable local feature enhancement module consists of an Encoder and a Decoder, and the entire module comprises multiple such units stacked together. Both the Encoder and Decoder share the same internal structure, illustrated in Fig. \ref{fig:coder}. Specifically, the feature map $X\in\mathbb{R}^{\mathcal{C}\times H\times W}$, derived from the multi-local feature extraction module, is transformed into a feature matrix $X\in\mathbb{R}^{HW\times C}$ through an embedding layer, which is subsequently used as the key and value in the Encoder's attention module. The learnable parameter $L_{i}$ serves as the query. Leveraging the powerful global modeling capabilities of the attention module, a connection between the learnable parameter $L_{i}$ and each pixel in the feature map is established, enabling the filtering of the most relevant local features for drowsiness detection. The Encoder produces the updated learnable parameter $L_{i+1}$.

We employ this updated learnable parameter $L_{i+1}$ as the key and value in the Decoder's attention module, with the feature matrix $X_{i}$ serving as the query. The attention mechanism in the Decoder enhances the detailed features within the feature map, resulting in the updated feature matrix $X_{i+1}$. In particular, we also apply attention dropout within multi-head attention, randomly setting the attention weights of one attention head to zero. More specifically, we first transform the learnable parameter $L_{i}$ and the feature matrix $X_{i}$ into query $Q$, key $K$, and value $V$ using the Eqs. (1)(2):
\begin{equation}
    Encoder\colon[Q,K,V]=\left[L_{i}W_{q},X_{i}W_{k},X_{i}W_{v}\right]
\end{equation}
\begin{equation}
    Decoder\colon[Q,K,V]=\left[X_{i}W_{q},L_{i}W_{k},L_{i}W_{v}\right]
\end{equation}
where$W_{q},W_{k}\in R^{c\times d_{k}},W_{v}\in R^{c\times d_{\nu}}$.

Next, the attention weights are calculated as shown in Eq. (3):
\begin{equation}
    A=Softmax\left(\frac{QK^{T}}{\sqrt{d_{k}}}\right)
\end{equation}

Finally, the output of the cross-attention layer is obtained through the weighted sum of the value $V$ calculated using the dropped attention weights, as shown in Eq. (4):
\begin{equation}
    O=\mathrm{Drop}(A)V
\end{equation}

The feedforward network layer consists of two fully connected layers and a GELU activation function. The output of the cross-attention layer is passed through two residual layers with normalization and a feedforward network layer to get the final output of the Encoder or Decoder.

We stack the Encoder and Decoder modules sequentially I times, updating the learnable patches $L_{i}$ and feature matrixs $X_{i}$ alternately. Ultimately, we obtain $X_{out}\in\mathbb{R}^{HW\times C}$, which incorporates both local and global feature information, making it more effective for drowsiness detection.

\section{Experiments}
In this section, we assess the performance of our proposed method on both our self-built drowsiness detection dataset and the RLDD dataset. We begin by introducing the datasets used and providing the experimental details. To establish the effectiveness of our method, we present the results of ablation experiments and compare our method with various established classification approaches and other drowsiness detection methods. Additionally, we offer some visualizations of the model results.

\subsection{Datasets}
To evaluate our proposed MAF drowsiness detection method, we used a self-built video dataset and the public dataset RLDD.

\begin{figure}[h]
    \centering
    \includegraphics[width=1\linewidth]{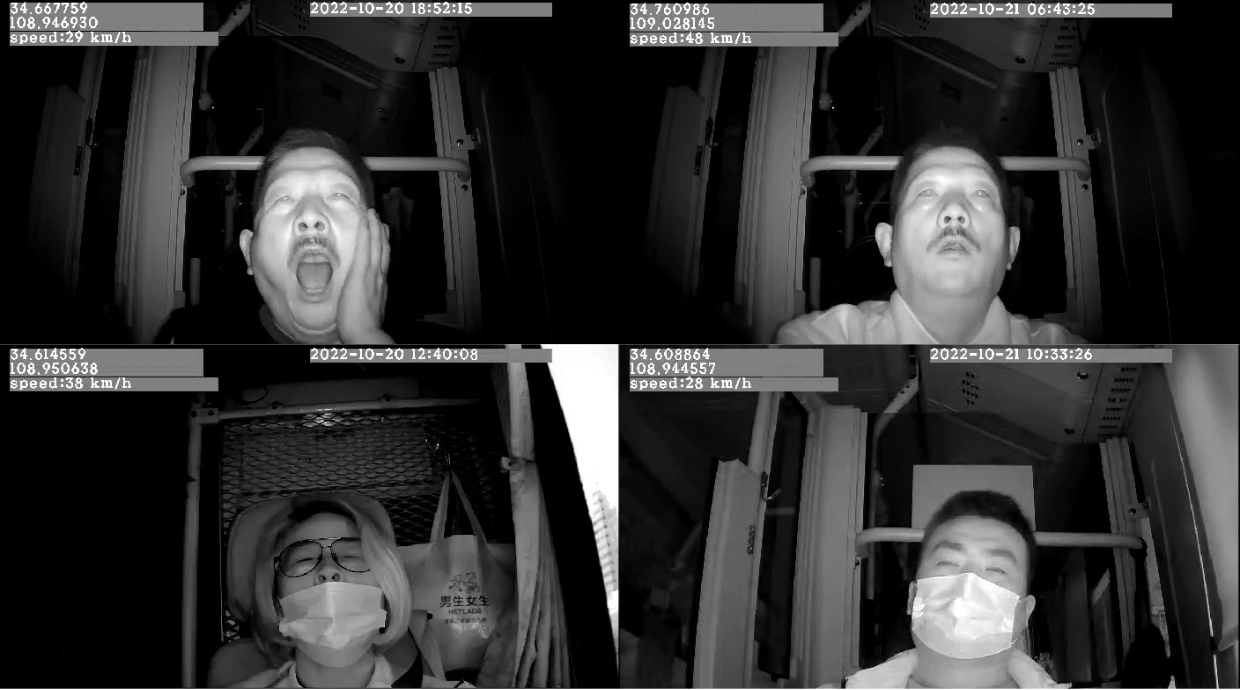}
    \caption{Examples of fatigue and non-fatigue states for day and night conditions in the self-built dataset}
    \label{fig:dataset}
\end{figure}

\textbf{Self-built Dataset:}Our self-built dataset comprises driver's face videos obtained from the surveillance systems of regularly operating buses in China, as well as simulated driving videos recorded by RGB cameras in stationary vehicles. The dataset includes content from 32 different drivers, including videos of real bus driver operations captured under daytime and nighttime conditions using a grayscale camera, as well as 142 simulated driving videos collected using an RGB camera. This dataset encompasses scenarios in which critical facial features, such as the mouth or eyes, are obscured, and conditions with low illumination. In Fig. \ref{fig:dataset}, you can view an example image from our dataset. We extracted images from the video data at intervals of 10 frames and labeled each frame as either 'drowsy' or 'non-drowsy.' Subsequently, we divided these images into a training dataset and a test dataset, maintaining a 9-to-1 ratio.

\begin{figure}[h]
    \centering
    \includegraphics[width=1\linewidth]{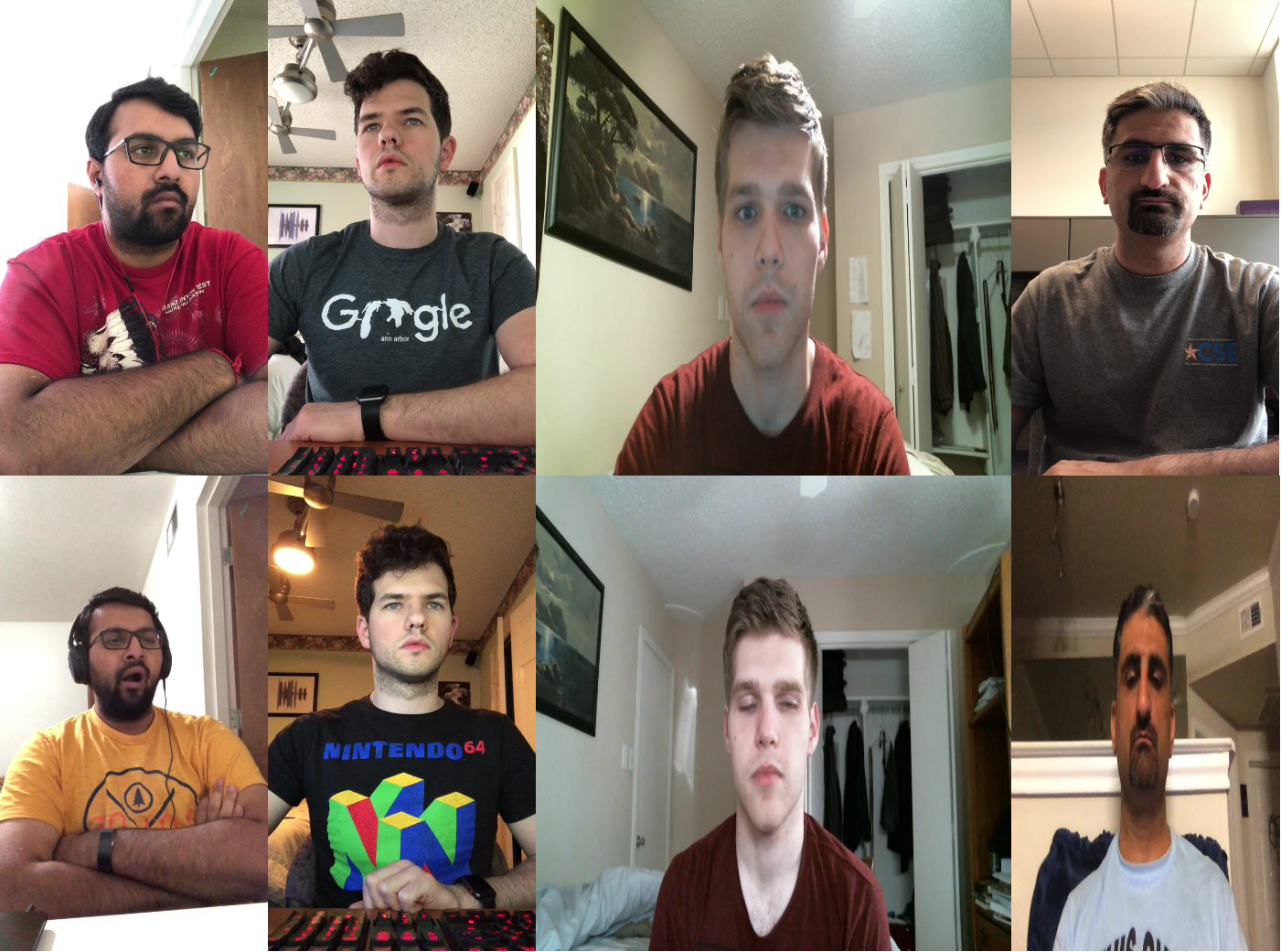}
    \caption{Examples of fatigue and non-fatigue states in the RLDD dataset}
    \label{fig:rldd}
\end{figure}

\textbf{RLDD:}The Real-Life Drowsiness Dataset (RLDD) comprises approximately 30 hours of RGB video footage from 60 healthy participants. Each participant contributed videos in three different categories: alert, hypovigilant, and drowsiness, resulting in a total of 180 videos. The participants, all aged 18 years or older, included 51 males and 9 females from diverse racial backgrounds (10 Caucasian, 5 non-white Hispanic, 30 Indo-Chinese Aryan and Dravidian, 8 Middle Eastern, and 7 East Asian) and a range of ages (20-59 years, with a mean age of 25 and a standard deviation of 6). For our analysis, we exclusively utilized a subset of 120 videos classified as 'alert' and 'drowsy' in the RLDD dataset. In Fig. \ref{fig:rldd}, you can see example images from the RLDD dataset. We segmented these videos into 10-second video sequences and meticulously reviewed the labeling of each segment. Similarly, we divided the dataset into a training dataset and a testing dataset, maintaining a 9-to-1 ratio.

\subsection{Implementation Details}
All models were implemented in Python using the PyTorch deep learning framework. In our experiments, we utilized the RTX 3090 GPU for both training and testing. The computer was equipped with an AMD Ryzen ThreadRipper Pro 3995WX CPU and 64GB of RAM, running on Ubuntu 22.04. For evaluating the test results, we employed the F1 score and accuracy (ACC) as our evaluation metrics.

In terms of implementation details, we utilized RetinaFace\cite{26deng2020retinaface} to extract facial regions from the original images, resizing all extracted face images to 112 × 112 pixels. As our backbone network, we employed IResNet50\cite{27deng2019arcface}. The number of Learnable Local Feature Enhancement Modules (I) was set to 2. We opted for stochastic gradient descent (SGD) as the optimizer with an initial learning rate of 1e-2, a momentum of 0.9, and a weight decay of 1e-5. The probabilities of attention dropout (p) were set to 0.6 and 0.4. The model was trained for 200 epochs with a batch size of 32, and we applied a cosine annealing learning rate decay strategy, adjusting the learning rate every 40 epochs. During training, the attention dropout layer was activated, while during prediction, the dropout layer was masked, and all attention maps were outputted.

\subsection{Comparative Experiment}

\begin{table}[h]
\centering
\begin{tabular}{ccc}
\hline
Models                    & F1 score       & ACC             \\ \hline
AlexNet\cite{28krizhevsky2012imagenet}           & 0.503          & 0.6914          \\
VGG-FaceNet\cite{29parkhi2015deep}       & 0              & 0.5508          \\
ResNet50\cite{30he2016deep}           & 0.911          & 0.9271          \\
Bhargava Reddy et al.\cite{34reddy2017real}     &0.918              & 0.9297            \\
DDD-IAA\cite{31park2016driver}           & 0.937          & 0.9514          \\
Mohit Dua et al.\cite{32dua2021deep} & 0.945          & 0.9549          \\
Syed Sameed Husain et al.\cite{33husain2022development} & 0.943          & 0.9492          \\ \hline
\textbf{Ours}             & \textbf{0.960} & \textbf{0.9688} \\ \hline
\end{tabular}
\caption{ACC and F1 scores of different methods on the self-
built dataset}
\label{tab:dataset}
\end{table}

We validated our models using the self-built dataset, training all models on the training dataset and subsequently validating them on the test dataset. We assessed both well-established classification models and models proposed in previous research using this dataset, reporting F1 scores and ACC for each model in Table \ref{tab:dataset}. Our proposed method outperforms classical classification models by an impressive margin, with accuracy improvements ranging from 1.9 to 41 percentage points. In contrast, VGG-FaceNet struggled to capture the hidden common features within the data, resulting in models that did not converge. Meanwhile, the model proposed by Bhargava Reddy and others, which utilized local images of the eyes and mouth, was susceptible to failure due to occlusions in these areas. Moreover, our method exhibits a 1.9 to 1.7 percentage point improvement compared to previous work, surpassing the detection model proposed by Mohit Dua et al., which also employs a ResNet50 backbone network. This underscores our method's superior capability in enabling the network to capture intricate features.

\begin{table}[h]
\centering
\begin{tabular}{ccc}
\hline
Models                    & F1 score        & ACC             \\ \hline
AlexNet\cite{28krizhevsky2012imagenet}              & 0               & 0.5142          \\
VGG-FaceNet\cite{29parkhi2015deep}               & 0               & 0.5142          \\
ResNet50\cite{30he2016deep}                  & 0.817           & 0.8149          \\
Bhargava Reddy et al.\cite{34reddy2017real}     &0.957                 &0.9591                 \\
DDD-IAA\cite{31park2016driver}                   & 0.835           & 0.8327          \\
Mohit Dua et al.\cite{32dua2021deep}          & 0.877           & 0.8701          \\
Syed Sameed Husain et al.\cite{33husain2022development} & 0.936          & 0.9377          \\ 
Jasper S. Wijnands et al.\cite{35wijnands2020real}  &0.9497    &0.9520 \\  \hline
\textbf{Ours}             & \textbf{0.991} & \textbf{0.9911} \\ \hline
\end{tabular}
\caption{ACC and F1 scores of different methods on the RLDD}
\label{tab:rldd}
\end{table}

Furthermore, we validated our proposed model on the RLDD dataset and reported the F1 scores and ACC in Table \ref{tab:rldd}. Notably, both AlexNet and VGG-FaceNet failed to capture the underlying common features in the data and were unable to converge. In comparison to detecting single-frame images in our self-built dataset, our proposed model demonstrated significant improvement in the detection of video sequences within the RLDD dataset compared to other models. When compared to ResNet50, our model exhibited a remarkable 18 percentage point improvement. Moreover, in contrast to the prior works, our model demonstrated an improvement ranging from 5.4 to 15.9 percentage points. The results of experiments on both our self-built dataset and the RLDD dataset underscore the effectiveness of our proposed model.

\subsection{Ablation Studies}

\begin{table*}[]
\centering
\begin{tabular}{ccccc}
\hline
Backbone &
  \begin{tabular}[c]{@{}c@{}}Learnable Local \\ Feature Enhancement Module\end{tabular} &
  \begin{tabular}[c]{@{}c@{}}Multi-local feature\\ extraction module\end{tabular} &
  Attention Drop &
  ACC \\ \hline
$\checkmark$ &   &   &   & 0.9286 \\
$\checkmark$ & $\checkmark$ &   &   & 0.9444 \\
$\checkmark$ &   & $\checkmark$ & $\checkmark$ & 0.9479 \\
$\checkmark$ & $\checkmark$ & $\checkmark$ &   & 0.9549 \\
$\checkmark$ & $\checkmark$ & $\checkmark$ & $\checkmark$ & 0.9688 \\ \hline
\end{tabular}
\caption{Results of ablation experiments on the self-constructed dataset, reporting ACC}
\label{tab:ablation}
\end{table*}

We conducted ablation experiments on the proposed model by separately removing modules from the MAF to demonstrate the importance of each module. The quantitative results of these ablation experiments are shown in Table \ref{tab:ablation}. In the absence of the attention dropout layer, the accuracy decreased by 1.4\%. When we removed either the learnable local feature enhancement module or the multi-local feature extraction module, the accuracy dropped by 2.1\% and 2.4\%, respectively. If all modules were removed, leaving only the backbone network, the accuracy rate decreased by 4\%.

\begin{table}[]
\centering
\begin{tabular}{cc}
\hline
\begin{tabular}[c]{@{}c@{}}The number of LANet\end{tabular} & ACC    \\ \hline
1                                                             & 0.6285 \\
2                                                             & 0.9688 \\
3                                                             & 0.9479 \\
4                                                             & 0.9479 \\ \hline
\end{tabular}
\caption{ACC of MAF with different number of LANet networks}
\label{tab:lanet}
\end{table}

In addition, we conducted further experiments to assess the impact of different numbers of LANet units on the model's performance. As indicated in Table \ref{tab:lanet}, when the number of LANet units is set to 1, there is a significant drop in performance. We attribute this drop to the fact that having too few LANet units causes the model to discard some features that are essential for classification. The best performance is achieved with 2 LANet units. However, as the number of LANet units continues to increase, there is a decrease in the model's performance. This decrease may be attributed to an excess of LANet units, leading the model to learn redundant and repetitive features, ultimately resulting in a performance decrease.

\begin{figure}[h]
    \centering
    \includegraphics[width=1\linewidth]{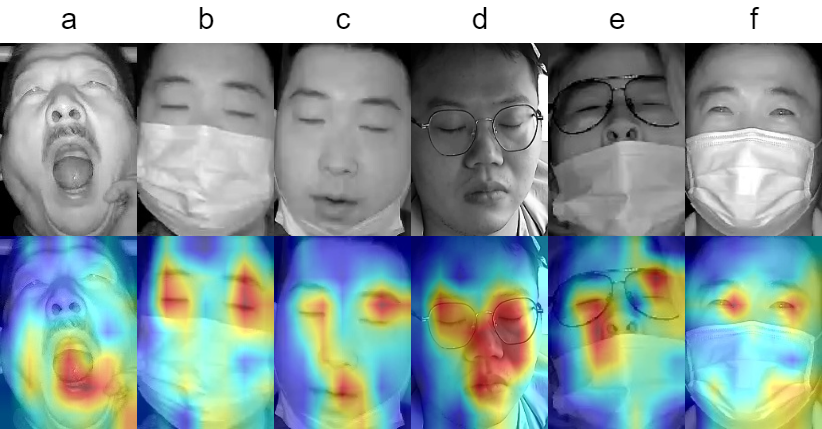}
    \caption{Final attention map visualization results of the Multi-Local Feature Extraction Module}
    \label{fig:att}
\end{figure}

\subsection{Attention Visualization}
To further investigate the effectiveness of our approach, we visualized the final attention graph of the multi-local feature extraction module. Specifically, we resized the attention map to match the input image's dimensions, normalized it to the [0, 255] range, converted the attention map to a heat map using OpenCV's COLORMAP\_JET method, and superimposed it onto the original image. In Fig. \ref{fig:att}, you can see the results of this attention visualization. The first row displays the original image, while the second row shows the image after applying the attention mapping. The results indicate that the model not only focuses on the most prominent facial features, such as the eyes and mouth, but also captures subtle changes in expressions around these regions. These subtle variations enable the model to better discern the driver's state. Furthermore, when the driver's face is partially covered by a mask, the model effectively avoids the masked area and captures relevant features from other facial regions.

\section{Conclusion}
In this paper, we introduce a Multi-Attention Fusion Drowsiness Detection Model (MAF) designed for the end-to-end learning of informative features and the determination of driver fatigue status based on input driver images. Our end-to-end detection method eliminates the need for complex, manually designed features and relies on straightforward category labeling of data, which not only conserves human resources but also facilitates easier deployment. Leveraging deep learning-based feature representation learning, our approach automates and streamlines the process of feature extraction, enabling us to categorize drivers into drowsy or non-drowsy states quickly and accurately. Experimental results on various datasets confirm the effectiveness of our model.

\printcredits

\bibliographystyle{cas-model2-names}

\bibliography{cas-refs}

\end{document}